\documentclass{article}
\pdfoutput=1

\usepackage[preprint,nonatbib]{neurips_2020}

\usepackage[utf8]{inputenc} 
\usepackage[T1]{fontenc}    
\usepackage{hyperref}       
\usepackage{url}            
\usepackage{booktabs}       
\usepackage{amsfonts}       
\usepackage{nicefrac}       
\usepackage{microtype}      

\usepackage[pdftex]{graphicx} 
\usepackage{wrapfig} 
\usepackage{siunitx} 
\usepackage{enumitem} 

\title{Dopant Network Processing Units: Towards Efficient Neural-network Emulators with High-capacity Nanoelectronic Nodes}

\author{Hans-Christian Ruiz Euler$^{1}$, Unai Alegre-Ibarra$^{1}$, Bram van de Ven$^{1}$, 
	\\ \textbf{Hajo Broersma$^{1,2}$, Peter A. Bobbert$^{1,3}$, Wilfred G. van der Wiel$^{1}$} \\
\texttt{\{h.ruiz, u.alegre, b.vandeven, h.j.broersma, p.a.bobbert, w.g.vanderwiel\}} \\ \texttt{@utwente.nl}
\\
$^{1}$MESA+ Institute \& BRAINS Center, $^{2}$Digital Society Institute, \\ University of Twente, NL;\\
$^{3}$Department of Applied Physics \& CCER, \\Eindhoven University of Technology, NL
}

\begin{document}

\maketitle

\begin{abstract}
The rapidly growing computational demands of deep neural networks require novel hardware designs. Recently, tunable nanoelectronic devices were developed based on hopping electrons through a network of dopant atoms in silicon. These “Dopant Network Processing Units” (DNPUs) are highly energy-efficient and have potentially very high throughput. By adapting the control voltages applied to its terminals, a single DNPU can solve a variety of linearly non-separable classification problems. However, using a single device has limitations due to the implicit single-node architecture. This paper presents a promising novel approach to neural information processing by introducing DNPUs as high-capacity neurons and moving from a single to a multi-neuron framework. By implementing and testing a small multi-DNPU classifier in hardware, we show that feed-forward DNPU networks improve the performance of a single DNPU from 77\% to 94\% test accuracy on a binary classification task with concentric classes on a plane. Furthermore, motivated by the integration of DNPUs with memristor arrays, we study the potential of using DNPUs in combination with linear layers. We show by simulation that a single-layer MNIST classifier with only 10 DNPUs achieves over 96\% test accuracy. Our results pave the road towards hardware neural-network emulators that offer atomic-scale information processing with low latency and energy consumption.
\end{abstract}

\section{Introduction}
\label{sec:introduction}

The success of deep neural networks (DNNs) comes with an exponential increase in the number of parameters and operations, which brings along high energy costs, high latency, and massive hardware infrastructure.
Moreover, due to a slowdown of Moore's law, the gap between the computational demands of DNNs and the efficiency of the hardware used to implement them is expected to grow.
There is a broad spectrum of research on hardware acceleration focused on obtaining state-of-the-art performance in DNNs, while reducing associated costs.
Solutions range from traditional approaches, which can be implemented on the short-term, to novel long-term approaches
trying to address fundamental problems such as that of the Von Neumann bottleneck or the slowdown of Moore's law \cite{xu2018scaling}.
\paragraph{State-of-the-art approaches.}
General-purpose hardware approaches for DNN acceleration typically employ a variety of temporal architectures to improve parallelism of multiply-accumulate (MAC) operations involved in convolutions and fully connected layers \cite{horowitz20141}. 
Specialised hardware approaches improve on the bottlenecks from the design of general computing. 
Since the energy consumption is dominated by the data movement during computation, these approaches are mostly focused on spatial architectures, reducing energy consumption by increasing the data reuse from low-cost memory hierarchy via optimized data-flows \cite{sze2017efficient}.
Specialised hardware encompasses FPGA-based acceleration \cite{guo2017survey}, ASIC-based acceleration \cite{qadeer2013convolution,chen2016diannao, han2016eie,zhang2016cambricon, albericio2016cnvlutin, jouppi2017datacenter}, or a combination of both \cite{nurvitadhi2018package}.
Furthermore, the development of specialised hardware enables DNN algorithm optimisation techniques to be jointly designed with the hardware \cite{vanhoucke2011improving,courbariaux2014training,han2015learning, han2015deep}.
 
\paragraph{Recent advances in neuromorphic computing.}
To reduce the impact of the data-movement bottleneck, some research aims at bringing memory closer to the computation, or even integrating the memory and the computation into a single architecture \cite{sze2017efficient}.
The latter approach encompasses the use of memristors, non-volatile electronic memory devices that can integrate MAC operations into the memory \cite{wong2015memory,ielmini2018memory}.
Recent developments show the potential of memristor arrays for implementing DNN connectivity fully in hardware \cite{yao2020fully}. Other research efforts try to overcome the hardware efficiency problem by taking a fundamentally different approach, such as in the case of spiking neural networks, which use sparse binary signals to compute asynchronously and in a massively parallel manner. 
Although these systems show high energy efficiency and low latency \cite{akopyan2015truenorth}, they tend not to support state-of-the-art DNN models, and their efficient training and proper benchmarking tends to be problematic \cite{sze2017efficient,pfeiffer2018deep}.

\paragraph{A promising novel technology.}
Recently, tunable nanoelectronic devices were developed capable of classifying linearly non-separable data, e.g. XOR \cite{chen2020classification,bose2015evolution}. These so-called dopant network devices are projected to have an energy efficiency in the order of 100 TOPS/Ws and a bandwidth of over 100 MHz, making them an attractive candidate for novel, unconventional hardware solutions for information processing.
Interestingly, the ability of these designless devices to perform the XOR operation coincides with that of individual human neocortical neurons, a recent observation that contradicts the conventional belief that this computation requires multilayered neural networks \cite{gidon2020dendritic}. 
Motivated by the above observations, we envision the usage of dopant network devices as \emph{high-capacity} nodes in a hardware architecture that emulates neural networks.
The implementation of these \emph{neural-network emulators} would have several advantages.
First, the expected small footprint, high throughput and energy efficiency would allow portability and low latency.
Second, massive parallelisation could be possible using many independent devices.
Third, computation is performed in-materio, i.e. by physical processes in the devices. Thus, the need of explicit arithmetic operations is greatly reduced, in particular if this technology is combined with memristors.
Moreover, in-materio computations could bypass the need of data management at the inference step because the learned parameters would be a fixed, physical characteristic of the system and computation would be reduced to physical processes transforming and propagating information.
Finally, high-capacity nodes may allow more compact neural network architectures that could bring additional benefits in terms of performance and efficiency. 

\paragraph{Towards novel neural-network emulators.}
In this paper, we take the first steps towards realising neural-network emulators with dopant network devices, giving new insights in its potential for this purpose. Section \ref{sec:dnpu} reviews the state-of-the-art of this nanotechnology, which we will henceforth call \emph{dopant network processing units} (DNPUs).
In Section \ref{sec:capacity}, we estimate the capacity of these complex, highly non-linear computational units in terms of an empirical estimate of the Vapnik–Chervonenkis dimension for binary classification of two-dimensional data. In order to study the viability of DNPU interconnection, Section \ref{sec:architectures} demonstrates how DNPU feed-forward network architectures can be designed, trained and implemented in hardware to solve classification tasks more accurately than a single device. In Section \ref{sec:mnist}, we explore novel, compact architectures that would reduce the number of parameters and/or operations, if inference were implemented with DNPU devices. We show, by simulation, that these high-capacity nodes allow for a single-layer classifier of hand-written digits from the MNIST dataset with high accuracy, using only 10 nodes. Section \ref{sec:discussion} discusses potential extensions of this work to large-scale neural-network emulators and their benefits.

\section{Dopant network processing unit}
\label{sec:dnpu}
\begin{figure}[hbtp!]
	\centering
	\vspace{-0.3cm}
	\includegraphics[width=0.8\linewidth]{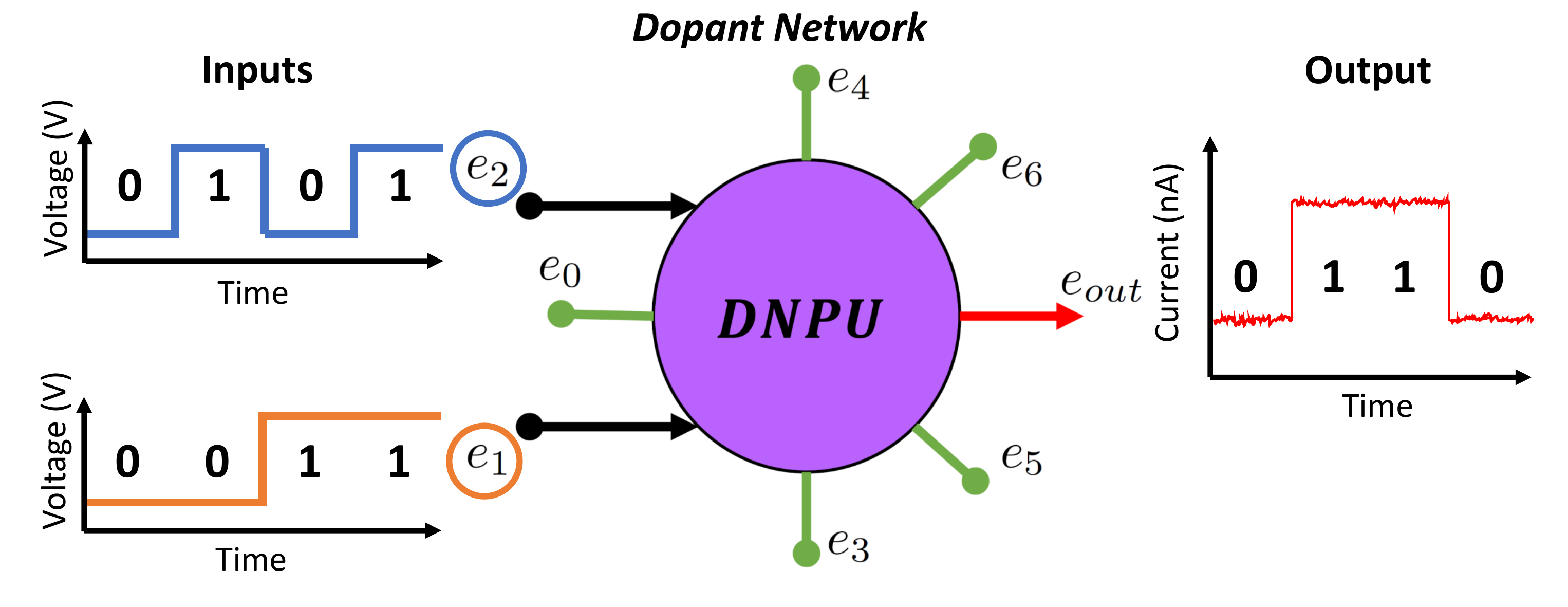}
	\caption{Sketch of a DNPU with eight electrodes \cite{chen2020classification}, where $e_{out}$ is the \textit{readout} electrode and the others can be either \textit{input} or \textit{control} electrodes, e.g. $e_{1,2}$ and $e_{0,3-6}$, respectively. To implement a classifier, voltages are applied to the input electrodes representing the features of the data, e.g. 0 and 1. Applying a learned voltage configuration to the control electrodes implements the classifier, e.g. XOR, as an output current representing the classes 0 and 1.}
	\label{fig:schematic}
	\vspace{-0.5cm}
\end{figure}

The basis of a DNPU is a lightly doped (n- or p-type) semiconductor with a nano-scale active region contacted by several electrodes, see Figure \ref{fig:schematic}. Different materials can be used as dopant or host and the number of electrodes can vary. 
Once we choose a readout electrode, the device can be activated by applying voltages to the remaining electrodes, which we call activation electrodes.
The dopants in the active region form an atomic-scale network through which the electrons can hop from one electrode to another.
This physical process results in an output current at the readout that depends non-linearly on the voltages applied at the activation electrodes. 
By tuning the voltages applied to some of the electrodes, the output current can be controlled as a function of the voltages at the remaining electrodes.
This tunability can be exploited to solve various linearly non-separable classification tasks \cite{chen2020classification, deep2020ruiz}.

For instance, let us assume that we want to create an XOR classifier using a DNPU \cite{chen2020classification}. For concreteness, let us consider a DNPU with eight electrodes, which are divided into seven activation electrodes $e_{0-6}$ and a single readout electrode $e_{out}$, as shown in Figure \ref{fig:schematic}. From the activation electrodes, $e_1$ and $e_2$ are chosen as data \textit{input electrodes}. These receive voltage-encoded signals representing the binary input features of the XOR classification task. We call these signals the \textit{input voltages}.
The remaining activation electrodes ($e_{0,3-6}$) are selected as \textit{control electrodes}, and are used to tune the relation between the input voltages and the output current. The voltage values applied at these electrodes are the learnable parameters and we call them \textit{control voltages}. Once the voltage values solving the task are found and applied to the corresponding control electrodes, inference is implemented by simply applying the input voltages to the input electrodes. 
Given the optimised control voltages, the functionality of the DNPU remains consistent over time. Therefore, even after switching to other voltage configurations or turning off the device, inference is still attained once we re-apply the optimised control voltages to solve that task \cite{chen2020classification}.\newline
Historically, DNPUs have been trained exploiting the concept of evolution-in-materio \cite{miller2014evolution}, which adopts a genetic algorithm to find adequate control voltages directly on-chip \cite{bose2015evolution, chen2020classification}.
A more recent approach \cite{deep2020ruiz} creates DNN surrogate models of DNPUs to predict the output current from the voltages applied to the activation electrodes. These surrogate models enable learning of the control voltages \emph{off-chip} by gradient descent (GD) using standard deep-learning packages. \newline
The off-chip GD training technique works as follows \cite{deep2020ruiz}. The input and output nodes of the DNN surrogate model correspond to the DNPU's activation and output electrodes, respectively. As before, we choose the DNN's inputs to represent the input and control electrodes. Then, we attach learnable \textit{control parameters} to the latter. The internal weights and biases of the model are henceforth kept frozen. Training the control parameters via GD exploits the fact that the gradient can be back-propagated to the input nodes with control parameters attached to them. 
Finally, the learned control parameters are validated by implementing the inference step on the device. This involves measuring the complete data set by arranging the samples sequentially in time, as shown in Figure \ref{fig:schematic}. Then, the input voltages representing each sample are applied one after another, while the control voltages remain fixed throughout the whole validation at the values found to solve the task.\newline
Throughout this paper, we use a boron-doped silicon (Si:B) DNPU with an active region of $300$ nm in diameter and the same configuration represented in Figure \ref{fig:schematic}. Furthermore, the device is trained using the off-chip training technique.
The DNPU surrogate model used for this consists of a feed-forward DNN with 5 hidden layers, each having 90 ReLU nodes. 
The data used to train the surrogate model consist of 4.5 million input-output pairs, sampled in a voltage range of [$-1.2$, $0.6$] V for electrodes $e_{0-4}$ and [$-0.7$, $0.3$] V for $e_{5,6}$, following the same procedure as in \cite{deep2020ruiz}. These working voltage ranges are determined by the electrical properties of the device. We split the data into training (90\%) and validation (10\%) sets.
The model was trained for 500 epochs in batches of 128 and a learning rate $\eta=0.0005$. Using an independently sampled test set of $4.5\times10^{5}$ samples, the root-mean-squared test error is found to be 1.4 nA, corresponding to 0.35\% of the total current output range between $-300$ and $100$ nA. The prediction error comes from measurement contributions and physical noise in the output current.
In the off-chip training, the corresponding control parameters are regularized with an L1-penalty outside of the working voltage ranges of the device. The penalty strength is set to $\alpha=1.0$ for all experiments.
We use PyTorch \cite{paszke2019pytorch} for both, training the DNN surrogate model and the off-chip training in all experiments. 
We use Adam \cite{kingma2014adam} with the default values for the moments and numerical stability parameters, unless explicitly mentioned otherwise.\newline
After off-chip training, validating the inference step on the device is performed as follows. The control voltages found are applied at the corresponding electrodes and kept fixed. Each data point is measured by applying its corresponding input voltages to the input electrodes for 0.8 seconds with a sampling frequency of 100 Hz. By measuring the output current for each data point long enough, we can account for the output noise. 

\section{Capacity of a single dopant network unit}
\label{sec:capacity}

\begin{figure}[hbtp!]
	\centering
	\includegraphics[width=1\linewidth]{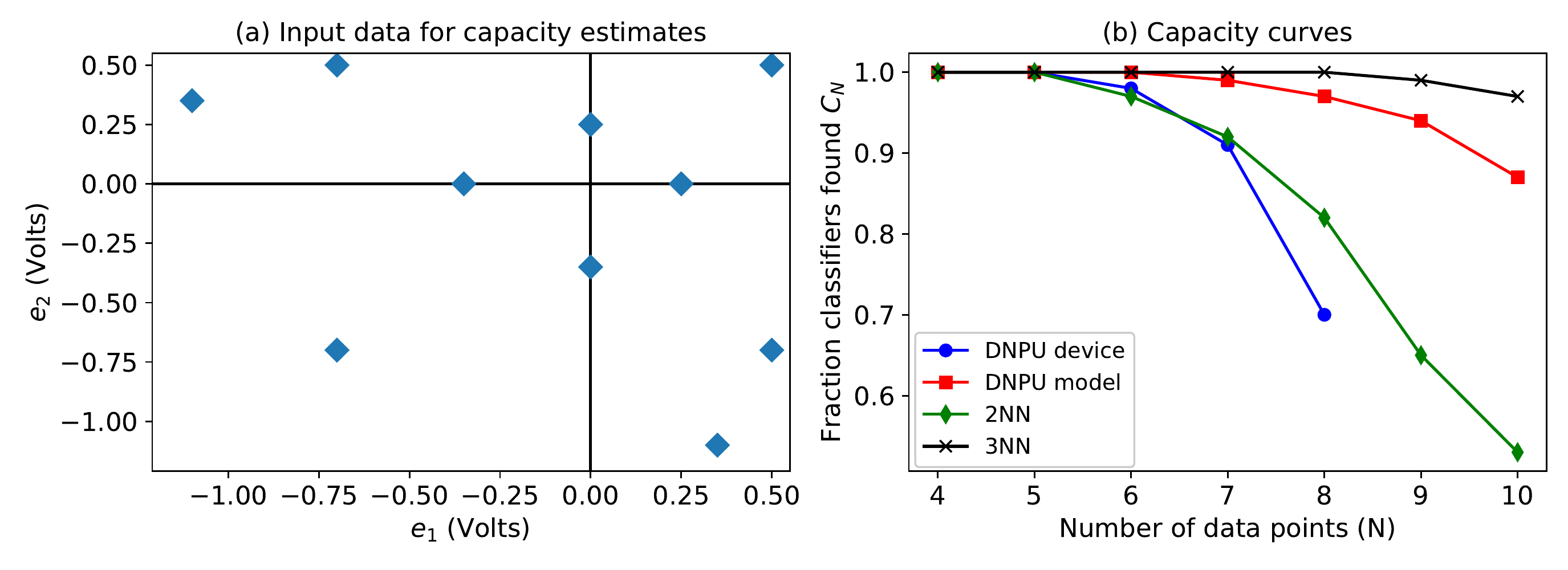}
	\caption{ (a) Input data for capacity estimates: \{(-0.7,-0.7), (-0.7,0.5), (0.5,-0.7), (0.5,0.5), (-0.35,0), (0.25,0), (0,-0.35), (0,0.25), (-1.1,0.35), (0.35,-1.1)\}. The classifiers should map these points to \{0,1\} in all possible ways. (b) Capacity of the DNPU surrogate model, the physical DNPU, and neural networks with a hidden layer of two (2NN) and  three (3NN) neurons.}
	\label{fig:vcdim}
\end{figure}

In this section, we show that a single DNPU has a computational capacity comparable to that of a small artificial neural network.
For this, we estimate the capacity of the DNPU surrogate model, the physical device and two small neural networks. 
As a measure of the computational capacity, we use an empirical estimate of the Vapnik-Chervonenkis (VC) dimension.
More specifically, given $N$ points, they can be mapped to $\{0,1\}$ in $2^N$ ways, i.e. there are $2^N$ possible classifiers.
Then, the capacity of a computational system for a given $N$ is defined as the fraction $C_N$ of classifiers that can be realized by the system.
Loosely speaking, the VC-dimension is defined as the largest $N$ such that $C_N=1$ \cite{friedman2001elements}.
By searching for all the classifiers, we can give an empirical estimate of the computational system's capacity in terms of the VC-dimension.\newline
We search for all classifiers on a fixed number $N$ of data points in the 2-dimensional plane, where $N=4,\dots,10$, see Figure \ref{fig:vcdim} (a). 
These points are chosen such that they fall within the working voltage range of the input electrodes of the DNPU ($e_1$ and $e_2$). The remaining electrodes $e_{0,3-6}$ are used as control electrodes.
Since XOR is the first non-trivial case, we start with $N=4$ and select the first four points of the list given in Figure \ref{fig:vcdim} (a). For $N=5$, we add the next point on the list and so on.\newline
Given $N$, we train each classifier using binary cross-entropy loss. This requires passing the output of the DNPU surrogate model through a \textit{decision node} consisting of a batch norm layer with a learnable affine transformation and a logistic function.
We train for 1,500 epochs with full batch, a learning rate $\eta=0.03$ and the Adam moment parameters $(\beta_{1},\beta_{2})=(0.995,0.999)$. The number of epochs is chosen to give all $2^N$ classifiers enough time to converge. However, no systematic hyper-parameter search was performed.
Furthermore, we allow for 15 attempts to find each classifier, which is considered to be found if its accuracy is $100\%$.\newline

Estimating the device's capacity requires special attention because of the noise in the current output. The off-chip training is the same as above but we add Gaussian noise to the model's output and reduce it after every attempt by a factor depending on the attempt's number $n$: $\sigma^{2}_{n+1}=(1-\frac{n+1}{15})\sigma^{2}_{n}$ with $\sigma^{2}_{n+1}$ the noise variance at attempt $n+1$ and $\sigma^{2}_{0}=1$.
This forces the training procedure to find robust solutions with a sufficient signal-to-noise ratio.\newline
The solutions obtained are validated on the physical device to determine if the classifier is found. In this case, a classifier is considered to be found if its accuracy is above a threshold $\theta = 1-0.5/N$, i.e. if more than $50\%$ of the measurements corresponding to one of the $N$ points can be classified correctly. 
As before, the predicted label is determined by the decision node which we re-train with the measured data, leaving out at random $20\%$ of the measurements as a test set to estimate the final accuracy on hardware. 
Once all classifiers are validated, we perform a new search for any failed cases and repeat this procedure one more time.\newline
The capacity of the DNPU is compared to that of two small neural networks with one hidden layer of 2 and 3 neurons.
Both networks have logistic activation functions and are trained with the same procedure as the DNPU’s surrogate model.
Figure \ref{fig:vcdim} shows the capacity estimates for all four systems.
The capacity of the physical device follows closely that of the neural network with two hidden nodes, both having a VC dimension of at least 5. The DNPU surrogate model has a VC dimension of at least 6 and the network with three hidden neurons has a VC dimension of at least 8.\newline 
As expected, we observe a steeper decay in the capacity of the device due to noise. 
However, the DNPU surrogate model has a much larger capacity than a network with two hidden neurons. This is consistent with the observation in \cite{deep2020ruiz} that the nano-electronic device can solve binary classification problems with closed decision boundaries, which requires conventionally at least three hidden neurons.
Hence, we can consider the DNPU capacity curves shown in Figure \ref{fig:vcdim}(b) as upper and lower empirical limits of their capacity.\newline
We want to remark at this point an important benefit of using the DNPU device for inference. Classification with the DNPU in this experiment requires only two arithmetic operations at the decision node but no explicit operation is required to compute the internal representation of the data that allows for linear separability. In contrast, the neural network with two hidden nodes requires at least 12 arithmetic operations without counting the contribution of non-linear activation functions. The difference is even more dramatic when compared to a network with three hidden nodes, which requires 18 operations. Interestingly, we also observe that we can solve a similar number of classification tasks with less parameters, namely 7 instead of the 9 (13) needed in the neural network with two (three) hidden nodes.

\section{Multi-DNPU Network}
\label{sec:architectures}
\begin{figure}[hbtp!]
	\centering
	\includegraphics[width=0.93\linewidth,trim=1.7cm 1.2cm 1cm 1cm]{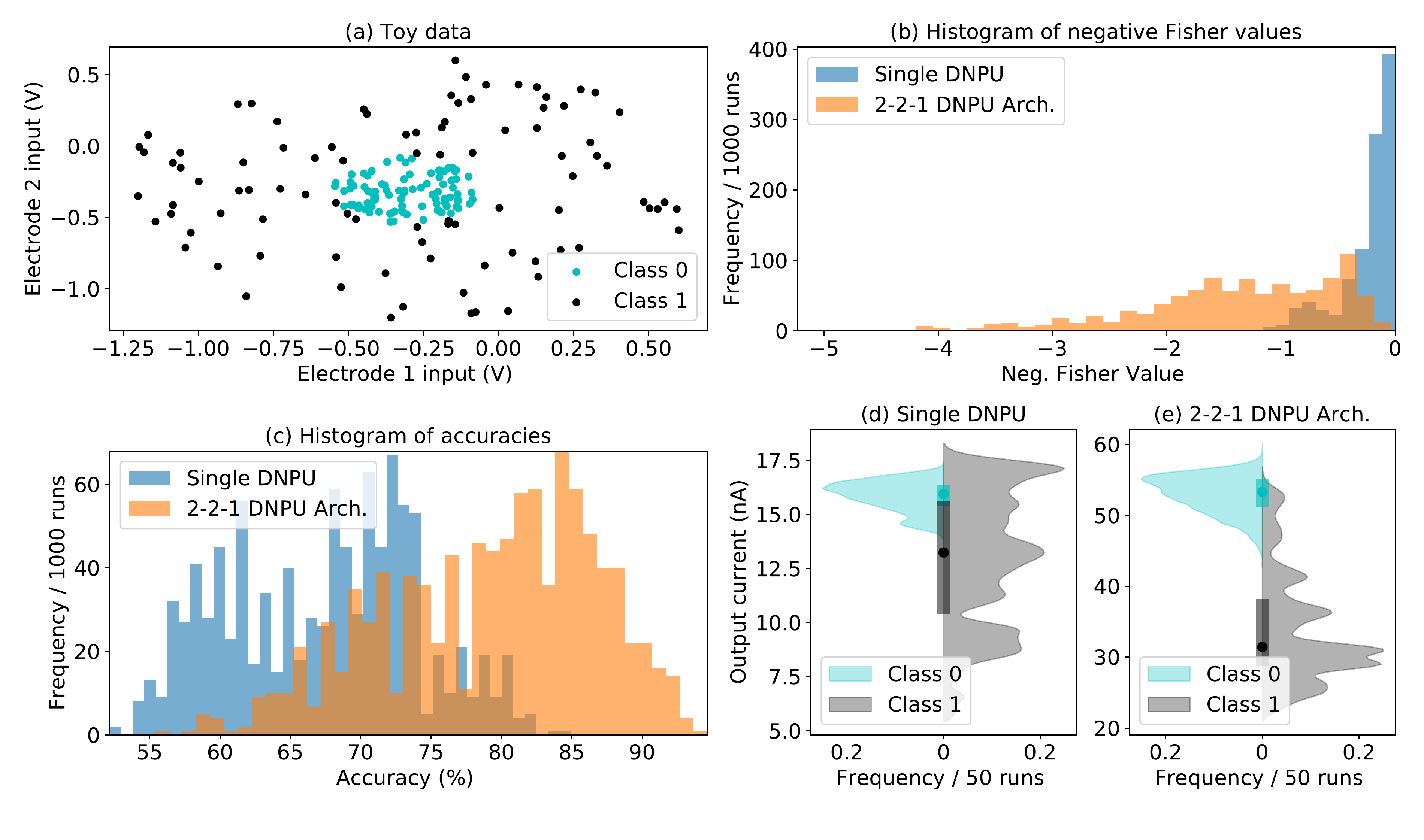}
	\caption{(a) Input data used in the experiment with a gap of 6.25 mV between the two classes. (b) Histogram of negative Fisher values for a single DNPU model compared to a 2-2-1 DNPU model architecture. (c) Histogram of accuracies for a single DNPU model compared to a 2-2-1 DNPU model architecture. (d) Outputs measured over 50 runs on a hardware DNPU using the control voltages obtained from the best result after off-chip training over 1,000 trials. The outputs are separated by classes on each side of the violin plot. (e) Outputs measured by time-multiplexing the hardware DNPU to implement the 2-2-1 architecture. The data is obtained with the same procedure as in (d).}
	\label{fig:all_figures}
\end{figure}

This section explores the potential of DNPU scalability by comparing the performance of a single DNPU device with a feed-forward network architecture of five DNPU devices.
Particularly, we consider a linearly non-separable binary classification problem consisting of two classes that fall into two concentric circles with a given separation \textit{gap}, see Figure \ref{fig:all_figures}(a).
In \cite{deep2020ruiz}, it is shown how a single DNPU is sufficient to resolve this classification problem with 100\% accuracy for a 0.4 V gap.
For the purpose of this section, the gap has been significantly reduced from 0.4 V to 6.25 mV, requiring a more accurate decision boundary in order to be solved.\newline
The data used consist of around 400 input points and their corresponding binary labels, equally divided into training and test sets.
Each class in each set has around 100 i.i.d. samples generated from random draws over a uniform distribution on their respective concentric circular areas, as shown in Figure \ref{fig:all_figures}(a). 
The same data have been used for both classifiers, the single DNPU and the DNPU network. \newline
All nodes in both classifiers have the same input, control and output electrode configurations, namely $e_1$ and $e_2$ for the input, $e_{0,3-6}$ for control and $e_{out}$ for the output, see Figure \ref{fig:schematic}.
The DNPU network consists of two input nodes, two hidden nodes and one output node. This 2-2-1 architecture is implemented on hardware by time-multiplexing the single DNPU to evaluate each node. 
That is, we mimic the 2-2-1 architecture using the same DNPU device but changing sequentially the control voltages corresponding to each node.
Each input unit receives a duplicate of the input data.
Then, each unit in the hidden layer receives the outputs of the input layer.
The output of each unit in the input and hidden layers is standardized and mapped linearly to the voltage range of the units in the next layer.
The last node receives the outputs of the two hidden units. 
This network has a total of 25 control parameters, distributed through the five electrodes $e_{0,2-6}$ in each node.\newline
To demonstrate the capabilities of the DNPU explicitly, we tackle training in two steps. 
First, we train the classifiers to separate the classes as much as possible using the negative of the Fisher's linear discriminant criterion as the loss function. 
This shows directly the capability of the DNPU systems to separate linearly non-separable classes. 
Second, once the data are separated in the output of the DNPU systems, the decision threshold for assigning the class is relegated to a simple logistic neuron.
The first training step uses 400 epochs with full batch updates and a learning rate of $\eta=0.0065$. There was no systematic hyper-parameter optimization. To account for the noise in the physical device, the classifiers are trained with Gaussian noise added to the output of each DNPU node and with a variance equal to the mean squared test error of the DNPU model, namely $\sigma^2=1.97$.
Finally, the logistic neuron is trained on the standardized output of the DNPU system and tested to estimate the accuracy of the classifier.

For each classifier, we run 1,000 off-chip training trials to compare the distribution of their solutions. Figure \ref{fig:all_figures}(b) shows the histogram of the negative Fisher values over all trials for both classifiers.
The best performance obtained for a single device is -1.16, while the 2-2-1 architecture obtains -5.23, which is around 4.5 times lower.
Figure \ref{fig:all_figures}(c) shows the accuracy histograms over all trials for both classifiers.
The highest accuracy in 1,000 trials for a single DNPU surrogate model is 84.95\%, while the best 2-2-1 architecture has an accuracy of 94.62\%, approximately ten percent points more.
Both results in Figures \ref{fig:all_figures} (b) and (c) show a significant improvement of the classification performance when using the multi-DNPU architecture in comparison to using a single DNPU.\newline
The best result of each classifier (minimum Fisher value and maximum accuracy over the 1,000 trials) is selected for validation on hardware.
Validation is performed for both training and test sets. 
The measurements for the 200 samples in each dataset gives approximately 16,000 current values.
Similar to the two-step procedure in the off-chip training, the device's output separates the input voltages of the training set into two distinct current levels. Then, to fine-tune the classifier on hardware, we re-train the logistic output neuron on the standardized current outputs.
To evaluate the performance on the test set, 50 validation trials are measured over the entire set for each classifier.  
All these measurements are then passed through the logistic neuron of the classifier to obtain the predicted labels.\newline
The validation of the single DNPU classifier on hardware yields an accuracy of 77.0\%  on the training set and an accuracy of 76.75\% on the test set.
Figure \ref{fig:all_figures}(d) shows the device's output distribution per class over the 50 validation trials using the test set. 
Since class 0 has a broad distribution that overlaps entirely with that of class 1, we observe a failure to separate the classes properly for a 6.25 mV gap. \newline
The validation of the 2-2-1 architecture gives an accuracy of 96.56\% on the training set and an accuracy of 93.98\% on the test set. Figure \ref{fig:all_figures}(e) shows the output current distribution per class over the 50 trials.
Contrary to the observations in Figure \ref{fig:all_figures}(d), the 2-2-1-architecture gives a good separation of the classes, i.e. their dominant modes in the output current distribution are well separated and have only some overlap at currents around 45 nA.\newline
The validation of both classifiers show consistent accuracy results. We estimated the variance of the classification accuracy on hardware over the 50 validation trials. We observe typical fluctuations in the accuracy of 1.05 percent points for both, the single DNPU's and the 2-2-1 DNPU architecture. These fluctuations are expected due to the noise in the output current. 

\section{Single-layer classifier for MNIST}
\label{sec:mnist}
\begin{figure}[hbtp!]
	\centering
	\vspace{-0.2cm}
	\includegraphics[width=1\linewidth,trim=3 1 1 1]{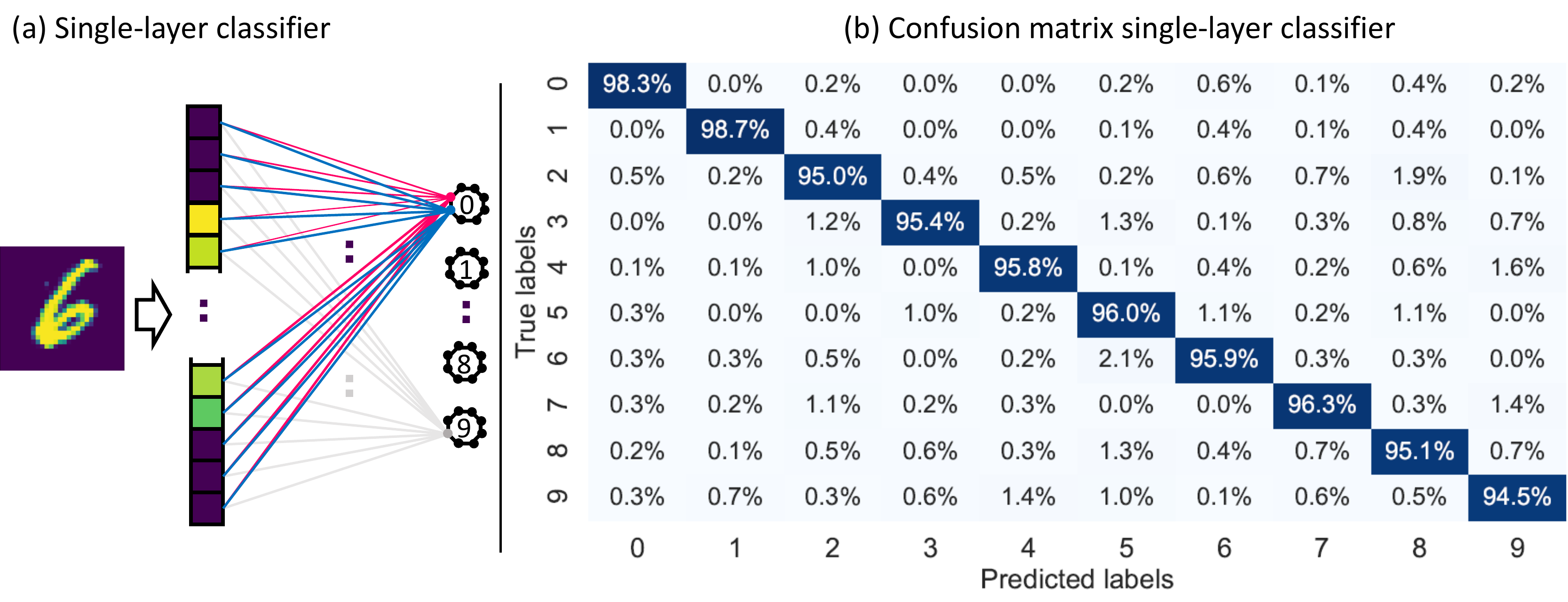}
	\caption{(a) Example architecture of a single-layer classifier. We use two distinct classifiers, one where each node receives the outputs of the linear layer at the 3 inputs corresponding to ($e_0,e_3,e_4$), and another with nodes receiving the linear outputs at all seven inputs. For clarity not all connections are shown. (b) Confusion matrix for MNIST classification of the single layer classifier with 7-dimensional local receptive fields.}
	\label{fig:cf_matrix}
	\vspace{-0.3cm}
\end{figure}

In this section we explore the potential of high-capacity neurons in combination with linear layers to process high-dimensional data efficiently. This approach is motivated by a possible integration of DNPUs with memristor arrays \cite{yao2020fully}.
We show by simulation how a single-layer network with only 10 high-capacity neurons can be used for classification on the MNIST data set, consisting of hand-written single digit (0-9) images of size $28\times28 = 784$ pixels in grey scale. There are 60,000 and 10,000 images in the training and test data, respectively. The former was split into 55,000 and 5,000 samples for training and validation, respectively.
Besides flattening the images, we standardize the grey scale of the pixels to values between -0.5 and 0.5. No other pre-processing is implemented.\newline
Our simulated DNPU network has a linear layer with $784$ input features, $30$ output features and no bias parameters. The outputs of the linear layer are fed into an output layer of $10$ DNPU surrogate models, each representing one class $\{0,\dots,9\}$, see Figure \ref{fig:cf_matrix}(a).
Each output unit has a local receptive field that receives three consecutive output features of the linear layer, i.e. the first three linear outputs feed into the first DNPU node, the next three into the second, and so on.
The input assignments to the DNPU models are the same for all nodes in the classifier, namely, the inputs corresponding to $e_0,e_3,e_4$ in Figure \ref{fig:schematic} are assigned as data inputs and the other inputs correspond to control parameters.\newline
The single-layer DNPU classifier is trained using cross-entropy loss.
We include weight decay for the linear layer parameters with a regularisation factor of $0.1$ to force its outputs to fall within the working voltage ranges of DNPUs, which are typically between $-1$ and $+1$ Volts.
We trained for 80 epochs with a learning rate $\eta=2\times10^{-5}$ and a mini-batch of 64 images.
The accuracy after training is 96.7\% (94.7\%) on the training (validation) set and 94.7\% on the test set.
There was no systematic hyper-parameter optimization but we observed that this result is fairly consistent across different values of the learning rate and mini-batch size. Nevertheless, a learning rate that is two orders of magnitude larger decreases performance significantly.\newline
As a benchmark, we compare our single-layer classifier to a neural network with a similar local receptive field architecture, i.e. there are 30 hidden neurons with ReLU activations fed in groups of three to the output layer consisting of 10 linear neurons. In this case however, we have bias parameters in the hidden and the output layers. Training this architecture is similar as above with the difference of a higher learning rate $\eta=3\times10^{-4}$.
This neural network achieves 94.5\% on the test data and 95.9\% (94.6\%) on the training (validation) set with a similar number of parameters to that of the single-layer DNPU classifier.\newline
Increasing the size of the local receptive fields of the DNPU nodes from three to seven ($e_{0-6}$) results in an accuracy of 96.1\% on the test set and 98.1\% (96.3\%) on training (validation) set.
Figure \ref{fig:cf_matrix}(b) shows the confusion matrix of this single layer classifier.
We observe that the typical "confusions" are also present here: the digit 9 is misclassified 2.4\% of the cases as the digits 4 and 5, and the digit 7 is misclassified 2.5\% of the cases as the digits 2 and 9.\newline
For comparison, the analogous neural network described before, with increased local receptive field size to seven, gives 96\% accuracy on the test set and 98.1\% (95.7\%) on training (validation) set, similar to the DNPU single-layer classifier.
Nevertheless, neglecting the non-linearity functions, this architecture must perform 14 arithmetic operations per output node to compute its activation, while a physical device would require a single evaluation per node. In addition, each output node requires 8 parameters, while the DNPU device with a 7-dimensional receptive field would not require any additional parameters.\newline
Hence, we can combine DNPUs with linear layers to solve high-dimensional classification problems with a comparable accuracy to standard neural networks that have a similar architecture but require a hidden layer. This motivates the integration of DNPU devices with memristors to realise neural-network emulators completely on hardware. 

\section{Discussion}
\label{sec:discussion}
\paragraph{Summary.}
Motivated by recent advances in nano-electronics, we have introduced tunable dopant network processing units (DNPUs) as promising candidates for nodes in hardware neural-network emulators, due to their high theoretical throughput and low energy consumption. We have demonstrated that a single DNPU has a capacity comparable to that of a small neural network with one hidden layer, while needing fewer learnable parameters. Furthermore, we have expanded previous work on a single DNPU to a multi-node framework by training and implementing a feed-forward DNPU network with 5 nodes. By time-multiplexing a single DNPU, this network achieves an accuracy of 94\% on a linearly non-separable binary classification task that is too challenging for a single DNPU. Finally, we have shown by simulation that DNPUs allow the classification of MNIST hand-written digits with over 96\% accuracy using a single-layer network with only 10 DNPUs. 
To the extent of our knowledge, the examples presented here are the first realisations of non-biological neural networks with high-capacity neurons.
\paragraph{Towards large-scale hardware neural-network emulators.}
Expanding on the examples presented in this paper, we envision the development of large-scale DNPU networks for efficient neural information processing, making use of the unique feature of DNPUs to perform non-linear computations that are commonly only performed in a neural network by combining arithmetic operations and non-linear activation functions. 
First, by leveraging the expected high throughput, virtual architectures can be implemented by time-multiplexing several independent DNPU devices in parallel. Second, integrating DNPUs with memristors would allow the co-allocation of memory and computation by implementing connectivity in-materio. Third, direct feed-forward interconnectivity of DNPUs in large circuits would provide a flexible neural network emulator. 
The advantage of this approach is that the learned control voltages can be applied directly to the hardware circuit for inference and, in principle, the same hardware can be deployed for various tasks by simply switching to different control voltage configurations. Furthermore, since computations in all three approaches take advantage of the physical properties of the materials, using DNPUs may reduce the parameters and operations required for inference. Finally, an important aspect of our approach for future research is the established off-chip training procedure, which makes research on neural network architectures with high-capacity neurons possible and allows the design of ad-hoc hardware solutions to emulate neural network functionality.

\bibliographystyle{plain}
\bibliography{postdoc}

\end{document}